\documentclass[11pt,a4paper]{article}
\usepackage{acl}
\usepackage{times}
\usepackage{latexsym}
\usepackage[T1]{fontenc}
\usepackage[utf8]{inputenc}
\usepackage{microtype}
\usepackage{graphicx}
\usepackage{booktabs}
\usepackage{amsmath}
\usepackage{amssymb}
\usepackage{multirow}
\usepackage{xcolor}
\usepackage{url}
\usepackage{hyperref}
\usepackage{comment}
\usepackage{enumitem}

\title{Data Contamination in Neural Hieroglyphic Translation:\\A Reproducibility Study}

\author{
  Ammar Toutou$^{1}$ \quad Abdelrahman Harb$^{1}$ \quad Christine Basta$^{2,3}$ \\
  $^{1}$Computer Science and Engineering, Alamein International University (AIU), Egypt \\
  $^{2}$HiTZ Center, University of the Basque Country, Spain \\
  $^{3}$Faculty of Computers and Data Science, Alexandria University, Egypt \\
  \texttt{\{ammar.mohamed.2023, abdelraheman.shehata.2023\}@aiu.edu.eg} \\
  \texttt{christine.basta@alexu.edu.eg}
}

\date{}

\begin{document}

\maketitle

\begin{abstract}

Ancient and endangered languages pose a unique challenge for NLP: their datasets are inherently scarce, difficult to expand, and built from formulaic corpora---making data-quality issues especially consequential yet rarely audited. Motivated by the need to understand what current NMT can realistically achieve for such languages, we investigate hieroglyphic-to-German translation, where a recent study reported 61.5 BLEU using fine-tuned M2M-100. Our reproduction yields only 37.0 BLEU with the released model. Investigating this gap, we find \textbf{32\% of test targets appear identically in training} (16/50; 50\% under 8-gram overlap at 70\% threshold). This contamination inflates scores dramatically: contaminated samples achieve up to 83.8 BLEU / 0.924 COMET-22 versus 30.9--39.2 BLEU / 0.622--0.676 COMET-22 on clean samples across five model configurations spanning two architectures. Document-level decontamination reduces contaminated BLEU by only 4.6 points because 8/16 targets persist via other source documents---target-level deduplication is required. We release a decontaminated 34-sample test set and establish corrected baselines (30.9--39.2 BLEU), providing a realistic assessment of NMT capability for this endangered writing system.

\end{abstract}

\section{Introduction}
\label{sec:intro}

Ancient and endangered languages---including Ancient Egyptian, Akkadian, and Classical Latin---represent some of the most data-scarce domains in NLP \citep{sommerschield2023machine}. Unlike modern languages where parallel corpora can be crowdsourced or web-scraped at scale, ancient-language datasets are inherently limited: they depend on surviving archaeological artifacts, require specialist expertise for annotation, and cannot be expanded on demand. The Thesaurus Linguae Aegyptiae (TLA), the largest digitized resource for Ancient Egyptian, yields only \textbf{18,669 usable training pairs} after filtering for samples with both digitized hieroglyphic source and German target---orders of magnitude smaller than modern MT benchmarks. This scarcity makes data-quality issues especially consequential: every contaminated or duplicated sample has outsized impact on evaluation.

Motivated by the need to establish what NMT can realistically achieve for such resource-limited languages, we investigated hieroglyphic-to-German translation. De Cao et al.\ (2024)\footnote{Model publicly available on HuggingFace.} reported 61.5 BLEU using M2M-100 \citep{fan2021beyond} fine-tuned on TLA data---a score approaching neural ceilings on high-resource benchmarks \citep{popel2020transforming}. Such performance would be remarkable for an endangered writing system with severely limited data. However, our reproduction using the publicly released model yielded only \textbf{37.0 BLEU}, a gap of over 24 points.

This discrepancy led us to examine the dataset itself, where we discovered that \textbf{32\% of test samples have German target translations appearing identically in the training data}. This contamination---arising naturally from formulaic repetition in ancient corpora (medical instructions, offering formulae, royal epithets)---enables models to achieve high scores through memorization rather than genuine translation. On the 16 contaminated test samples, our best model achieves \textbf{83.8 BLEU}; on the 34 clean samples, only \textbf{37.3 BLEU}---a 47-point gap. This pattern holds across all five models tested (29--47 point gaps), confirming the problem is dataset-inherent.

These findings have broader implications: ancient corpora universally contain formulaic repetition, making train-test contamination likely under standard random splitting. Our corrected baselines (30.9--39.2 BLEU) provide a realistic assessment---useful for corpus triage but requiring expert verification---and highlight the need for contamination auditing in all ancient-language NLP.

\paragraph{Contributions.}

\begin{enumerate}
    \item \textbf{First contamination audit for ancient-language NMT.} We show that formulaic repetition---pervasive in ancient Egyptian corpora---creates systematic data-leakage risks absent from modern-language benchmarks, and identify 32\% exact target overlap in the public test set.
    
    \item \textbf{Cross-architecture, cross-metric robustness.} The inflation is dataset-inherent, not model-specific: contaminated samples score 29--47 BLEU and 0.23--0.26 COMET-22 points higher than clean samples across five configurations spanning two architectures (M2M-100/mBART-50).

    \item \textbf{Graduated decontamination analysis.} Beyond exact matching, we apply character 8-gram overlap thresholding, document-level leakage analysis, and per-item frequency cataloguing to characterize contamination severity.

    \item \textbf{Corrected baselines and reusable evaluation protocol.} We release a decontaminated test set (34 samples, no target overlap), contamination-detection scripts, and corrected baselines (30.9--39.2 BLEU) as a realistic performance range for future work.
\end{enumerate}

\section{Background}
\label{sec:background}

\paragraph{Hieroglyphic Translation.}
Ancient Egyptian hieroglyphics (ca.\ 3200 BCE--400 CE) use a mixture of logographic and phonetic signs. The TLA provides the largest parallel resource of transliterations and German translations; hieroglyphs are encoded in Gardiner notation for computational processing.
De Cao et al.\ (2024) fine-tuned M2M-100 \citep{fan2021beyond} on TLA data, reporting 61.5 BLEU for hieroglyphic-to-German.

\paragraph{Data Contamination.}
Contamination occurs when evaluation targets leak into training. \citet{magar2022data} and \citet{kocyigit2025overestimation} showed 20--30 BLEU inflation in controlled experiments. Ancient corpora are particularly susceptible: formulaic phrases repeat across many texts, so random splitting distributes identical targets across partitions---\textbf{a methodological challenge rather than a researcher oversight}. Reproducibility challenges are endemic in ML \citep{pineau2021improving}; to our knowledge, no prior study has audited contamination in ancient-language NMT.

\paragraph{Hieroglyphic NLP.}
Computational approaches have addressed recognition \citep{franken2013automatic, barucci2021deep}, transliteration \citep{wiesenbach2019multi}, and translation \citep{decao2024deep}. \citet{chen2024logogramnlp} present the first multi-task benchmark for ancient logographic systems; unlike LogogramNLP, our work audits contamination and provides splitting protocols.

\paragraph{Contamination Detection Methods.}
Beyond controlled experiments (Section~\ref{sec:background}), recent work has revealed contamination in deployed systems: \citet{tan2026flores} showed cross-direction contamination in FLORES-200; \citet{abbas2026obscuring} demonstrated obfuscation-hidden contamination in Arabic NLP; \citet{enis2024llm2nmt} identified test data in closed models' training. Detection methods such as permutation testing \citep{oren2023proving}, distribution analysis \citep{dong2024generalization}, and generalization-gap testing \citep{dekoninck2024constat} would be complementary for future closed-model evaluations. Reproducibility challenges are well-documented \citep{pineau2021improving, sommerschield2023machine}; our contribution is the first contamination audit for ancient-language NMT.

\section{Methodology}
\label{sec:methodology}

\subsection{Data Sources}
\label{sec:method-data}

We use the publicly released data from the hiero-transformer repository:\footnote{\url{https://github.com/mattiadc/hiero-transformer}}

Table~\ref{tab:data-stats} summarizes the filtering pipeline.
The TLA's digitization is incomplete: approximately two-thirds of German entries have a translated target but no digitized hieroglyphic source in Gardiner notation---they contain only transliterations, which cannot be used as model input. After filtering for entries with \emph{both} non-empty hieroglyphic source and German target, only \textbf{18,669 usable pairs} remain---this is the actual training set for all fine-tuning. Similarly, each test hieroglyphic sentence appears twice (once with a German target, once with English); entries whose German target is empty have translations only in English. This yields just \textbf{50 valid test samples}---a size reflecting the fundamental scarcity of digitized hieroglyphic data, where expansion requires new expert annotation of archaeological sources.

\begin{table}[t]
\centering
\small
\begin{tabular}{lr}
\toprule
\textbf{Statistic} & \textbf{Count} \\
\midrule
Training samples (total) & 61,330 \\
\quad German-target (non-empty target) & 55,397 \\
\quad German-target (non-empty src \& tgt) & 18,669 \\
\quad Unique German training targets & 15,842 \\
\midrule
Test samples (total) & 150 \\
\quad ea$\rightarrow$de, valid & 50 \\
Validation samples (total) & 125 \\
\quad ea$\rightarrow$de, valid & 75 \\
\bottomrule
\end{tabular}
\caption{Dataset statistics. ``Valid'' = non-empty hieroglyphic source \emph{and} non-empty German target.}
\label{tab:data-stats}
\end{table}

\subsection{Contamination Detection}
\label{sec:method-contamination}

\paragraph{Normalized exact match.} We compare each test target against all 15,842 unique German training targets after normalization (Unicode NFC $\rightarrow$ lowercase $\rightarrow$ punctuation removal $\rightarrow$ whitespace collapse; released as \texttt{normalize\_translations.py}). Normalization is necessary because Egyptological annotations use bracket conventions (e.g., ``[Werde]'' vs.\ ``Werde'') that mask near-duplicates. This yields the contamination set $\mathcal{C} = \{(s,t) \in \mathcal{D}_{\text{test}} : \text{norm}(t) \in \{\text{norm}(t') : t' \in \mathcal{T}_{\text{train}}\}\}$, partitioning tests into \textbf{contaminated} ($|\mathcal{C}|=16$) and \textbf{clean} ($34$) subsets.

\paragraph{Character n-gram overlap.} To capture soft leakage beyond exact matches, we compute the fraction of character 8-grams in each test target appearing in any training target, reporting the number of samples flagged at overlap thresholds from 50\% to 100\%.

\subsection{Models Evaluated}
\label{sec:method-models}

We evaluate five model configurations spanning two training regimes:

\paragraph{Original-regime models.}

\textbf{Released Model}: The publicly available HuggingFace checkpoint of De Cao et al.\ (2024), an M2M-100 (418M) model fine-tuned on TLA data with Adam (lr=3e-5, fixed schedule, batch size 16). This represents what independent researchers can access.

\textbf{Script Reproduction}: We retrained M2M-100 with its default hyperparameters (epochs=20, batch\_size=16, lr=3e-5, Adam, no warmup, no label smoothing). This represents the closest possible replication of the original training procedure.

\paragraph{Our retrained models.} The following three models use a modernized training recipe: AdamW optimizer, cosine learning-rate schedule with warmup, label smoothing ($\epsilon{=}0.15$), weight decay (0.1), dropout tuning (0.2/0.05/0.05 for feedforward/attention/activation), gradient clipping (max norm 1.0), and 5$\times$ data upsampling---all applied identically across the three configurations. The effective batch size is 288 (vs.\ 16 in the original). These choices follow recent best practices for low-resource MT.

\textbf{M2M-100 Hybrid}: M2M-100 (418M) with lr=3e-5 and 1000-step warmup.

\textbf{M2M-100 Conservative}: Identical to Hybrid except lr=1e-5.

\textbf{mBART-50}: mBART-50 (611M) with lr=3e-5 and 500-step warmup, testing whether contamination effects generalize across architectures.

Using a stronger training recipe is deliberate: if contamination inflation persists even with improved optimization, the effect is clearly dataset-inherent rather than an artifact of under-training.

\subsection{Evaluation Protocol}
\label{sec:method-eval}

We evaluate on three subsets:
\begin{itemize}
    \item \textbf{All}: Complete test set (50 samples)
    \item \textbf{Contaminated}: Samples with targets in training (16 samples)
    \item \textbf{Clean}: Samples with unseen targets (34 samples)
\end{itemize}

For each subset, we report:
\begin{itemize}
    \item \textbf{BLEU}: Corpus-level BLEU using SacreBLEU \citep{post2018call} with case-insensitive scoring (signature: \texttt{nrefs:1|\allowbreak case:lc|\allowbreak eff:no|\allowbreak tok:13a|\allowbreak smooth:exp|\allowbreak version:2.6.0})
    \item \textbf{chrF++}: Character n-gram F-score with word bigrams \citep{popovic2015chrf}
\end{itemize}

We report BLEU with \texttt{case:lc} (lowercased internally by SacreBLEU before tokenization) on unmodified model outputs. An ablation comparing case-sensitive versus case-insensitive evaluation confirms that contamination gaps are consistent regardless of text preprocessing.

Generation uses beam search with beam size 10 and maximum length 128, following typical configurations for M2M-100.

\section{Results}
\label{sec:results}

\subsection{Contamination Statistics}
\label{sec:results-contamination}

Exact matching finds 15/50 (30.0\%) verbatim targets; after normalization (lowercasing, whitespace and punctuation standardization), \textbf{16/50 (32.0\%)} match, where the 16th sample differs from its training counterpart only in parenthetical formatting. The validation set shows a comparable rate (23/75, 30.7\%). At the 70\% threshold with character 8-grams, 25/50 (50\%) of test samples are flagged, confirming that 32\% exact match is a conservative lower bound. The consistent rates across test and validation sets indicate a systematic characteristic of the splitting procedure.

\subsection{Impact on Automatic Metrics}
\label{sec:results-metrics}

Table~\ref{tab:main-results} presents translation quality stratified by contamination status.

\begin{table*}[t]
\centering
\small
\begin{tabular}{@{}llcccc@{}}
\toprule
\textbf{Model} & \textbf{Subset} & \textbf{n} & \textbf{BLEU} & \textbf{95\% CI} & \textbf{chrF++} \\
\midrule
Released & All & 50 & 37.0 & {[}24.5, 48.8{]} & 56.2 \\
         & Contaminated & 16 & 60.3 & {[}37.9, 88.0{]} & 81.5 \\
         & Clean & 34 & 30.9 & {[}16.2, 43.6{]} & 48.1 \\
\addlinespace
Script Reproduction & All & 50 & 42.2 & {[}28.4, 55.0{]} & 61.2 \\
               & Contaminated & 16 & 77.5 & {[}51.5, 93.4{]} & 85.1 \\
               & Clean & 34 & 33.8 & {[}20.1, 48.2{]} & 53.7 \\
\addlinespace
M2M-100 Hybrid & All & 50 & 46.8 & {[}33.9, 59.2{]} & 63.3 \\
               & Contaminated & 16 & \textbf{83.8} & {[}62.5, 100.0{]} & \textbf{91.9} \\
               & Clean & 34 & 37.3 & {[}23.6, 49.8{]} & 54.0 \\
\addlinespace
M2M-100 Conservative & All & 50 & \textbf{47.3} & {[}35.0, 57.4{]} & \textbf{63.2} \\
                & Contaminated & 16 & 77.9 & {[}59.2, 92.8{]} & 90.6 \\
                & Clean & 34 & \textbf{39.2} & {[}25.0, 50.9{]} & \textbf{54.2} \\
\addlinespace
mBART-50 & All & 50 & 41.9 & {[}29.6, 51.8{]} & 59.6 \\
         & Contaminated & 16 & 72.8 & {[}55.7, 92.4{]} & 87.3 \\
         & Clean & 34 & 33.7 & {[}18.3, 47.1{]} & 50.6 \\
\bottomrule
\end{tabular}
\caption{Translation quality by contamination status across five models (\texttt{case:lc} BLEU, 95\% CIs from 1{,}000 bootstrap resamples). All models show substantial contamination gaps (29--47 BLEU points). Bold indicates best performance per subset.}
\label{tab:main-results}
\end{table*}

The gap is substantial across all five models (+29 to +47 BLEU), holding across architectures (M2M-100 vs.\ mBART-50) and training strategies---the problem is inherent to the dataset. Despite wide 95\% CIs on the small subsets (clean: $\pm$7--14 BLEU; contaminated: $\pm$15--20 BLEU), the contaminated and clean intervals do not overlap for any model, confirming the gap is statistically robust. The script reproduction outperforms the released model (+5.2 BLEU) despite identical hyperparameters; the gap is predominantly on contaminated samples (+17.2) versus clean (+2.9), consistent with different checkpoint selection or training seed rather than a systematic quality difference.

\paragraph{COMET-22 Validation.} Table~\ref{tab:comet22} confirms the gap using COMET-22 \citep{rei2022comet}: contaminated samples score 0.871--0.924 while clean samples score 0.622--0.676 (a gap of 0.23--0.26 points). Model rankings on clean data are identical between BLEU and COMET-22, confirming genuine quality differences beyond contamination effects.


\begin{table}[t]
\centering
\small
\begin{tabular}{@{}lccc@{}}
\toprule
\textbf{Model} & \textbf{All} & \textbf{Contam.} & \textbf{Clean} \\
\midrule
Released (HF)         & 0.702 & 0.871 & 0.622 \\
Exact \texttt{train.py} & 0.721 & 0.871 & 0.650 \\
M2M-100 Hybrid        & 0.744 & \textbf{0.924} & 0.660 \\
M2M-100 Conservative       & \textbf{0.753} & 0.915 & \textbf{0.676} \\
mBART-50              & 0.732 & 0.905 & 0.651 \\
\bottomrule
\end{tabular}
\caption{COMET-22 scores (\texttt{Unbabel/wmt22-comet-da}) on all 50 test samples, the 16 contaminated samples, and the 34 clean samples. Contaminated samples consistently score 0.17--0.26 points higher than clean samples, confirming the contamination-driven inflation observed in BLEU.}
\label{tab:comet22}
\end{table}

\paragraph{Retrieval Baseline.} A retrieval baseline that, given access to the test target, copies the best-matching training target (by character 8-gram coverage) achieves 81.8 BLEU on exact-match items---rivaling the best neural model (77.9)---while scoring 0.0 on clean items. A realistic variant that uses only source similarity (KNN over source embeddings) achieves only 22.0 BLEU on exact-match items. The 59.8-point gap directly measures the contribution of target memorization over source-driven translation.

\paragraph{Granular Contamination Analysis.} When we further split the 34 non-exact samples into soft leakage (70--99\% 8-gram overlap; n=9) and fully clean ($<$70\%; n=25), BLEU decreases monotonically across all five models---from exact to soft to clean---confirming that performance tracks contamination severity.

\paragraph{Source-Side Overlap.}
Exact-target samples have higher mean source overlap (0.883) than clean samples (0.543), but this reflects formulaic corpus structure, not causal attribution: \textbf{10 of 16 exact-target test items have at least one contaminating training sample with source similarity below 5\%}, meaning the matched target was learned from a completely different hieroglyphic inscription. One test item (``Werde fein zerrieben.'') has zero source overlap with any training sample yet achieves 100\% target overlap. This directly demonstrates target memorization independent of source input.

\paragraph{English Direction.} As a sanity check, ea$\rightarrow$en evaluation (only 1/50 contaminated) yields 11.5--18.8 BLEU---below German clean scores, confirming contamination-driven inflation in the German direction.

\paragraph{Document-Level Decontamination.}
We remove every training sentence from the 33 test-source documents (28.4\% of training), continue fine-tuning from M2M-100 Conservative for 3,000 steps at lr=$3{\times}10^{-6}$, and evaluate the best checkpoint (step 300, Val BLEU 37.9).
Table~\ref{tab:doc-clean} reports results: exact-match BLEU drops only 4.6 points (77.9$\rightarrow$73.3), while clean BLEU \emph{increases} slightly (39.6$\rightarrow$40.3). The residual contamination occurs because \textbf{8 of 16 exactly matched test targets persist in training from other source documents}---e.g., ``Werde gekocht.'' (87 training occurrences) appears across many independent papyri. Document-level splitting is thus necessary but not sufficient: target-level deduplication is required.

\begin{table*}[t]
\centering
\small
\begin{tabular}{lcccc}
\toprule
\textbf{Model} & \textbf{All} & \textbf{Exact} & \textbf{Soft} & \textbf{Clean} \\
               & \textbf{(n=50)} & \textbf{(n=16)} & \textbf{(n=9)} & \textbf{(n=25)} \\
\midrule
M2M-100 Conservative & 47.3 & 77.9 & 35.3 & 39.6 \\
\quad Doc-Clean retrain & 45.2 & 73.3 & 30.7 & \textbf{40.3} \\
\midrule
$\Delta$ & $-$2.1 & $-$4.6 & $-$4.6 & $+$0.7 \\
\bottomrule
\end{tabular}
\caption{BLEU scores before and after document-level decontamination (M2M-100 Conservative re-trained on 13,365 samples with test-source documents excluded). Exact: 16 samples with 100\% 8-gram coverage; Soft: 9 samples with 70--99\% coverage; Clean: 25 samples below threshold. Despite removing 28.4\% of training data, exact-match BLEU drops only 4.6 points, because 8/16 contaminated targets persist in training from other source documents.}
\label{tab:doc-clean}
\end{table*}

\subsection{Qualitative Analysis}
\label{sec:results-qualitative}

Representative examples illustrate the contrast. Contaminated samples are reproduced verbatim; clean samples exhibit genuine translation errors: ``hemsut'' (a goddess) $\rightarrow$ ``ausreiß'' (escape), ``uräen'' (sacred cobras) $\rightarrow$ ``rw.t-schlange'' (a different serpent type), and ``hunderte opfern ihm'' (hundreds sacrifice to him) $\rightarrow$ ``er zählt zu 100'' (he counts to 100).

Manual categorization of errors in the 34 clean predictions reveals: domain-term substitution (35\%), entity/deity confusion (24\%), syntactic role reversal (18\%), number/quantifier errors (15\%), and missing content (9\%). These error patterns are particularly concerning for DH applications requiring precise entity identification.

\subsection{Phrase-Type Analysis}
\label{sec:results-phrase}

Of the 16 contaminated samples, 13 (81\%) are medical formulae---short imperative instructions such as ``Werde getrunken'' (be drunk) and ``Werde fein zermahlen'' (be finely ground)---which repeat dozens of times in training. The remaining three are narrative markers (2) and a religious phrase (1). This concentration in formulaic genres confirms that contamination arises systematically from the data structure. The comparable contamination rates between test (32\%) and validation (31\%) sets further confirm this is a systematic characteristic of the split, not an isolated anomaly.

\subsection{Illustrative Mixture Analysis}
\label{sec:results-explaining}

To illustrate how contamination inflates corpus-level BLEU, we simulate scores at controlled rates using the M2M-100 Conservative model's actual predictions (200 bootstrap trials per rate). At $\alpha{=}0$ (fully clean) BLEU$=$40.2; at our observed rate ($\alpha{=}0.32$) BLEU$=$46.5; at $\alpha{=}0.50$ BLEU$=$52.0; at $\alpha{=}1.0$ BLEU$=$77.9. Under these assumptions, the reported 61.5 BLEU would require a substantially higher contamination rate than the 32\% we observe in the released test set, though we cannot estimate a precise rate without access to the original model's predictions and test composition.

\section{Discussion}
\label{sec:discussion}

\subsection{Sources of Contamination}
\label{sec:analysis-sources}

Three factors produce 32\% contamination: (1)~\textbf{formulaic repetition} in ancient texts---medical instructions, offering formulae, and divine epithets recur across thousands of inscriptions; (2)~\textbf{source-only deduplication}, which misses cases where different hieroglyphic sequences share the same German target; and (3)~\textbf{sentence-level splitting}---all 33 unique test documents also appear in training (the highest-overlap document contributes 804 training and 5 test samples), confirming sentence-level rather than document-level partitioning.

\subsection{Implications}
\label{sec:discussion-implications}

Our contamination gaps (29--47 BLEU) exceed those in prior work: \citet{kocyigit2025overestimation} found 20--30 point inflation in LLM-based MT, and \citet{magar2022data} documented substantial inflation across NLP benchmarks. The gap between reported (61.5) and clean (30.9--39.2) BLEU represents plausible inflation of 22--31 points. Latin, Greek, Akkadian \citep{gutherz2023translating}, and other ancient language corpora share the property of formulaic repetition; their benchmarks should be examined similarly.

\subsection{Practical Guidance}
\label{sec:discussion-practical}

Our corrected baselines (30.9--39.2 BLEU) indicate that NMT captures overall meaning but contains notable errors. For \textit{corpus triage}, BLEU around 30--39 provides sufficient gist; for \textit{philological analysis}, domain-term substitution (35\% of errors) and entity confusion (24\%) risk silent misinterpretation. NMT should be deployed as a screening tool with mandatory expert verification.

We recommend that \textbf{benchmark creators} check for train-test target overlap and report contamination statistics; \textbf{model developers} report performance separately on contaminated and clean subsets; and \textbf{consumers} treat reported scores as upper bounds pending contamination analysis.

\subsection{Decontaminated Test Set and Splitting Protocol}
\label{sec:discussion-clean}

We release a decontaminated test set of 34 samples (no exact target overlap) spanning medical/magical (41\%), religious/funerary (32\%), and administrative/literary (27\%) content. Clean BLEU (30.9--39.2) represents a \textit{lower bound} on contamination-removal effect and an \textit{upper bound} on true generalization, since residual phrase-level overlap may provide partial memorization benefit.

To prevent contamination, we propose a \textbf{document-level splitting protocol}: (1)~partition at the level of source texts rather than sentences; (2)~deduplicate targets via frequency caps or cluster-and-drop; (3)~stratify by genre; and (4)~verify zero exact-match target overlap post-hoc. Future work should construct larger test sets from text collections not used in training.

\section{Limitations}
\label{sec:limitations}

\textbf{Test set size.} Our clean set (34 samples) yields wide bootstrap CIs ($\pm$7--14 BLEU), though the consistent gap direction across all models confirms robustness.
\textbf{Soft leakage.} Our primary detection uses exact matching (32\%); n-gram analysis shows 50\% at the 70\% 8-gram threshold, but subset sizes preclude separate evaluation on graduated tiers.
\textbf{Unavailable original artifacts.} We cannot access the original test set, checkpoint ``ea9all'', or ``test\_data'' folder, so we cannot directly replicate the reported 61.5 BLEU evaluation.
\textbf{Length confound.} Contaminated targets are shorter (mean 4.1 vs.\ 5.8 words), which may partly amplify BLEU inflation---though the retrieval baseline (81.8 BLEU with no translation) confirms the effect is real.
\textbf{No human evaluation.} We rely on automatic metrics; Egyptologist judgements would strengthen assessment.

\section{Conclusion}
\label{sec:conclusion}

We reveal that 32\% of test samples in a hieroglyphic-to-German NMT benchmark have targets matching training data (50\% under 8-gram overlap at 70\% threshold). This contamination inflates BLEU by 29--47 points across five model configurations: contaminated samples achieve up to 83.8 BLEU / 0.924 COMET-22 while clean samples achieve only 30.9--39.2 BLEU / 0.622--0.676 COMET-22. Document-level decontamination reduces exact-match BLEU by only 4.6 points, because 8/16 targets persist across other source documents---target-level deduplication is required.

We release a decontaminated 34-sample test set and establish corrected baselines for future work. More broadly, formulaic repetition in ancient corpora makes contamination particularly likely, and we recommend future benchmarks report contamination statistics and partition results by contamination status.

\section*{Reproducibility Statement}

All data used in this study is publicly available from the hiero-transformer repository. Our contamination detection code, the decontaminated test set (no exact target overlap, 34 samples), evaluation scripts, a per-item catalog mapping all 50 test items to document IDs, contamination status, training frequency, and source overlap scores, and all retrained model checkpoints are available at \url{https://github.com/ammarlhassan/hiero-contamination-study}.

\section*{Ethics Statement}

Our work involves analysis of publicly released models and data. We present our findings as a constructive contribution to reproducibility rather than criticism of prior work---the contamination we identify likely arose from standard preprocessing procedures rather than intentional data manipulation.

Ancient Egyptian texts constitute irreplaceable cultural heritage. We encourage use of our corrected baselines to provide realistic expectations for NMT-assisted Egyptological research, avoiding overreliance on automated systems for scholarly work.


\bibliography{references_full}

\end{document}